\documentclass[letterpaper]{article} 
\usepackage{aaai24}  
\usepackage{times}  
\usepackage{helvet}  
\usepackage{courier}  
\usepackage[hyphens]{url}  
\usepackage{graphicx} 
\usepackage{natbib}  
\usepackage{caption} 
\usepackage{algorithm}
\usepackage{algorithmic}

\usepackage{newfloat}
\usepackage{listings}
\DeclareCaptionStyle{ruled}{labelfont=normalfont,labelsep=colon,strut=off} 
\floatstyle{ruled}
\newfloat{listing}{tb}{lst}{}
\floatname{listing}{Listing}
\usepackage{subcaption}
\usepackage{booktabs}
\usepackage{multirow}
\usepackage{makecell}

\title{COMMA: Co-Articulated Multi-Modal Learning}
\author {
    Lianyu Hu\textsuperscript{\rm 1},
    Liqing Gao\textsuperscript{\rm 1},
    Zekang Liu\textsuperscript{\rm 1},
    Chi-Man Pun\textsuperscript{\rm 2},
    Wei Feng\textsuperscript{\rm 1}\thanks{Corresponding author}
}
\affiliations {
    \textsuperscript{\rm 1}College of Intelligence and Computing, Tianjin University, China\\
    \textsuperscript{\rm 2}Department of Computer and Information Science, University of Macau, China\\
    hly2021,lqgao,lzk100953@tju.edu.cn,cmpun@umac.mo,wfeng@ieee.org
}

\usepackage{bibentry}

\begin{document}

\maketitle

\begin{abstract}
Pretrained large-scale vision-language models such as CLIP have demonstrated excellent generalizability over a series of downstream tasks. However, they are sensitive to the variation of input text prompts and need a selection of prompt templates to achieve satisfactory performance. Recently, various methods have been proposed to dynamically learn the prompts as the textual inputs to avoid the requirements of laboring hand-crafted prompt engineering in the fine-tuning process. We notice that these methods are suboptimal in two aspects. First, the prompts of the vision and language branches in these methods are usually separated or uni-directionally correlated. Thus, the prompts of both branches are not fully correlated and may not provide enough guidance to align the representations of both branches. Second, it's observed that most previous methods usually achieve better performance on seen classes but cause performance degeneration on unseen classes compared to CLIP. This is because the essential generic knowledge learned in the pretraining stage is partly forgotten in the fine-tuning process. In this paper, we propose Co-Articulated Multi-Modal Learning (COMMA) to handle the above limitations. Especially, our method considers prompts from both branches to generate the prompts to enhance the representation alignment of both branches. Besides, to alleviate forgetting about the essential knowledge, we minimize the feature discrepancy between the learned prompts and the embeddings of hand-crafted prompts in the pre-trained CLIP in the late transformer layers. We evaluate our method across three representative tasks of generalization to novel classes, new target datasets and unseen domain shifts. Experimental results demonstrate the superiority of our method by exhibiting a favorable performance boost upon all tasks with high efficiency. Code is available at \url{https://github.com/hulianyuyy/COMMA}
\end{abstract}

\section{Introduction}
The increase of web data with aligned large-scale text-image pairs has greatly facilitated the development of foundation vision-language models (VLMs) such as CLIP~\cite{radford2021learning}. Thanks to the supervision provided by the natural language, these models have demonstrated excellent generalization performance over a series of downstream tasks and could reason about open-vocabulary visual concepts~\cite{gao2021clip,fang2021clip2video,cheng2021improving}. During inference, a set of hand-crafted prompts such as 'a photo of [category]' is used as a query for the text encoder. The output text embeddings are matched with the visual embeddings generated by the image encoder to predict the output class. 

Despite the impressive generalizability of CLIP over novel scenarios, its massive model scale and requirements of training data make it infeasible to fine-tune the full model in the downstream tasks. Fine-tuning the whole model also easily forgets the beneficial knowledge acquired in the training stage and overfits the downstream data. To handle the above limitations, a series of works~\cite{radford2021learning,jin2021good} are dedicated to designing better hand-crafted prompts to fit downstream tasks. However, hand-crafted prompts require careful selections with intensive labors, which may also be suboptimal in depicting the characteristics of novel scenarios. Recently, many methods~\cite{shu2022test,zhou2022conditional,zhou2022learning}  propose to treat the prompts as textual embeddings and update them in the fine-tuning process to better coordinate with the VLMs. In this procedure, only the learnable prompts are updated and the original parameters of VLMs are fixed, which greatly reduces the requirements of computations.

We argue that these approaches still own two major drawbacks. First, the prompts of the vision and language branches in these methods are usually separated or uni-directionally correlated (the vision branch is uni-directionally influenced by the text branch only). As the goal of VLMs is to better match the embeddings of vision and language branches, disjointed vision and language prompts may hinder modelling the correlation of output embeddings in two branches. Second, it has been observed that most previous methods usually achieve superior performance on seen classes but demonstrate worse generalizability on unseen classes compared to CLIP. This is because the essential generic knowledge acquired in the pretraining process is partly forgotten in the fine-tuning procedure. 

\begin{figure}[t]
    \centering
    \includegraphics[width=\linewidth]{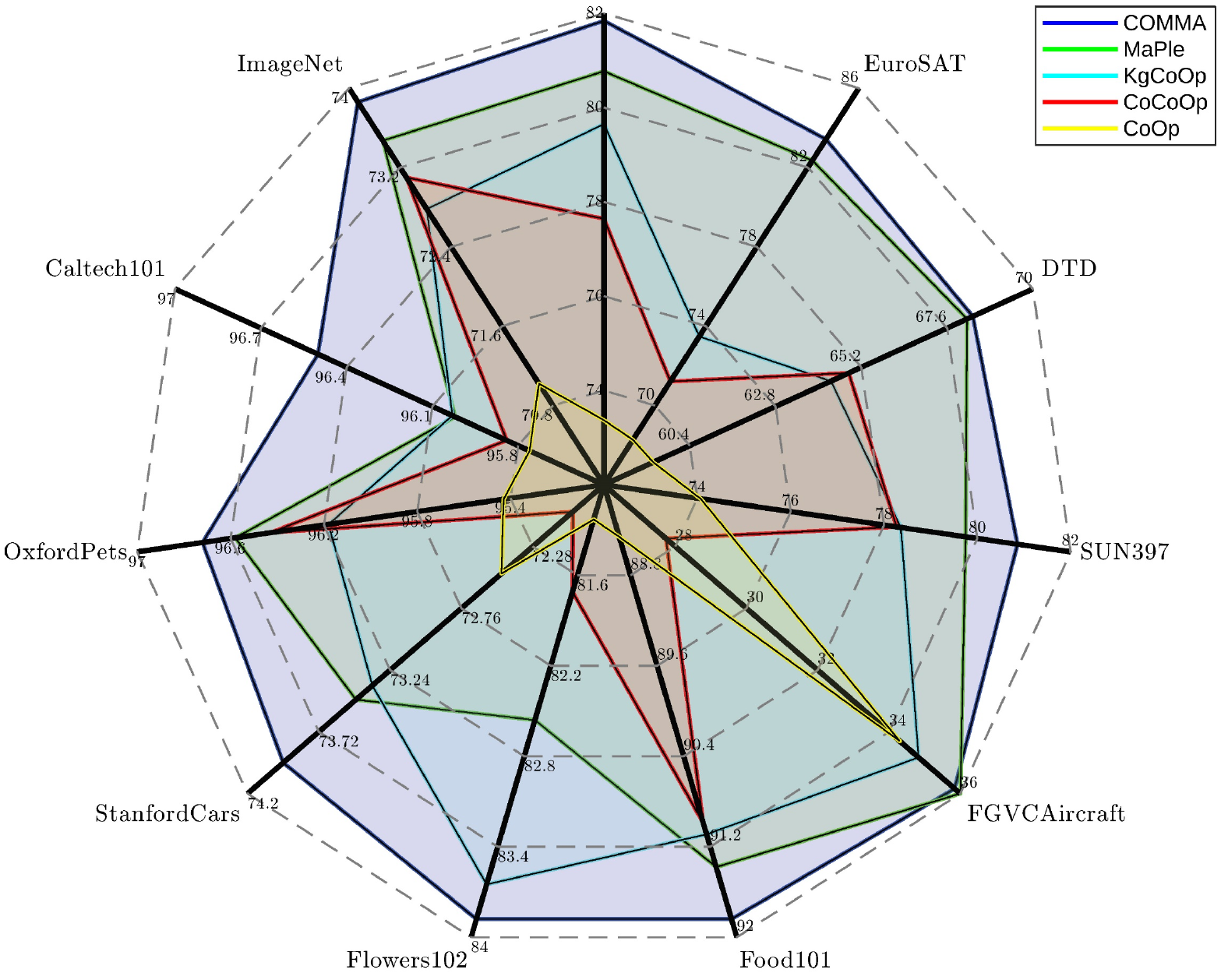} 
    \caption{COMMA outperforms state-of-the-art methods across 10/11 diverse image recognition datasets on the base-to-novel generalization task.}
    \label{fig1}
  \end{figure}

To handle the above limitations, we propose Co-Articulated Multi-Modal Learning (COMMA) in this paper. Especially, to enhance the correlations of prompts in both branches, we generate prompts of the next layer based on preceding prompts in both branches. In this case, the prompt embeddings of both branches are well correlated and could provide enough guidance for the next layer to align representations of both branches. Besides, to alleviate forgetting about the essential knowledge acquired in large-scale training data, we try to minimize the discrepancy between the learned prompts and the embeddings of hand-crafted prompts in the pretrained CLIP. The generic knowledge can be better preserved and adapted to novel classes in the fine-tuning stage. Our extensive experiments on three key representative settings including base-to-novel generalization, cross-dataset evaluation, and domain generalization demonstrate the strength of COMMA. Especially, on base-to-novel generalization, our method outperforms other approaches across 10/11 datasets as shown in fig.~\ref{fig1}. Further, our COMMA also demonstrates excellent generalizability over all datasets in the cross-dataset transfer and domain generalization settings, achieving consistent performance boost. Thanks to its streamlined design, COMMA exhibits improved training and inference efficiency compared to previous methods.

\section{Related Work}
\subsection{Vision Language Models}
Recently, the equipment of large-scale image-text pairs has greatly facilitated the development of Vision Language Models (VLMs). Previous methods usually adopt region-based~\cite{anderson2018bottom} or grid-based~\cite{jiang2020defense,nguyen2020movie} approaches to model the correlations between vision and language. However, the internal relations between two modalities are not fully captured by such a design. Recently, a series of models like CLIP~\cite{radford2021learning}, ALIGN~\cite{jia2021scaling}, FLIP~\cite{yao2021filip} and BLIP~\cite{li2022blip} are introduced to capture the correlations between image and text in a contrastive manner. They learn joint image-language representation by maximizing the similarity of positive pairs and pushing away those negative pairs. Aided by the supervision of natural language, they have demonstrated impressive performance over a broad series of downstream tasks. However, their massive model size and requirement of training data limit their applications in resource-constrained downstream tasks. How to better exhibit their potential in those novel concepts with high efficiency is still a challenging problem. Many works have demonstrated better performance on downstream tasks by using tailored methods to adapt VLMs for few-shot image-recognition~\cite{zhang2021tip,sung2022vl}, object detection~\cite{gu2021open,feng2022promptdet,maaz2022class}, and segmentation~\cite{ding2022decoupling,luddecke2022image}. 

\subsection{Prompt Learning}
Large language models often require instructions in the form of sentences, known as text prompts, to better understand the task. These prompts can be hand-crafted~\cite{jin2021good} or automatically learned~\cite{houlsby2019parameter,liu2023pre} during the fine-tuning process, while the latter is referred to as prompt learning. The trend has first appeared in the natural language processing (NLP) field where some methods~\cite{lester2021power,li2021prefix,liu2021p} propose to prepend a series of learnable prompts to the inputs or the intermediate features to adapt the learned representations to new tasks. A similar tendency has also arisen in the visual domain~\cite{jia2022visual,wang2022dualprompt} and vision-language domain~\cite{gao2021clip,khattak2023maple,yao2023visual}. While most previous methods separately consider prompts in multi-modal branches, we try to enhance their correlations to guide the alignment of representations in both branches. 

\subsection{Prompt Learning in Vision Language Models}
Inspired by prompt learning in NLP, many methods propose to adapt VLMs by learning the prompt tokens through end-to-end fine-tuning. The original parameters of VLMs are fixed and only a few extra learnable prompt parameters are updated in this procedure. CoOp~\cite{zhou2022learning} first replaces the hand-crafted prompts with learnable soft prompts in the first layer to adapt to VLMs. CoCoOp~\cite{zhou2022conditional} proposes to generate an image-conditional prompt to utilize the power of input features. ProGrad~\cite{zhu2022prompt} only updates the prompts whose gradient is aligned to the “general knowledge” generated by the original prompts. KgCoOp~\cite{yao2023visual} tries to align the output embeddings of the text encoder with those of the pretrained CLIP to preserve beneficial information. MaPLe~\cite{khattak2023maple} learns soft prompts in both vision and language branches to better align their representations. We note that the prompts in these methods are usually separated, which hinders adopting multi-modal information to better align the representations of both branches. Besides, these methods usually cause performance degeneration over unseen classes compared to CLIP, with much worse generalizability. Our work is the first to explore gathering beneficial information from both branches to better guide the prompt generation process to well align multi-modal representations.

\section{Method}
Our method focuses on how to improve the generalization performance of a large-scale VLM over a series of downstream tasks. To alleviate overfitting the downstream tasks and avoid incurring huge training computational costs, the parameters of both image encoder and text encoder in the original VLM are kept fixed, while only the parameters of the prompts are updated in the fine-tuning process. To better demonstrate the effects of our COMMA, we first give a brief review of VLMs by taking CLIP~\cite{radford2021learning} as an example, and then recap typical visual prompt learning methods like CoOp and MaPLe to derivative our method. 

\subsection{Preliminaries}
We build our model based on a pre-trained VLM, CLIP, which consists of a text and vision encoder. CLIP encodes an image $I\in \mathcal{R}^{H\times W\times 3}$ and a concurrent text description to match their output embeddings. We follow previous methods to use a vision transformer (ViT)~\cite{dosovitskiy2020image} based CLIP model. 

\textbf{Encoding Image:} An image $I\in \mathcal{R}^{H\times W\times 3}$ is first split into $M$ patches with equal intervals, and then reshaped and projected into patch embeddings $E_0\in \mathcal{R}^{M\times d_v}$. These patch embeddings are then sent into a $K$-layer transformer $\mathcal{V}$ along with a learnable class token (CLS) $c_i$. The calculation process of each transformer layer can be represented as:
\begin{equation}
\label{e1}
      [c_i, E_i] = \mathcal{V}_i([c_{i-1}, E_{i-1}]), i\in [0,\dots, K-1].
\end{equation} 
To obtain a final representation for the input image $I$, the CLS token $c_{K-1}$ of the last transformer layer is extracted and projected to the common V-L latent embedding space via a projection function $p_{v}$ as :
\begin{equation}
\label{e2}
    x = p_{v}(c_K).
    \end{equation} 

\textbf{Encoding Text:} the input text descriptions are tokenized and then projected into word embeddings $W_0=[w_0^1, w_0^2,\cdots,w_0^N]\in \mathcal{R}^{N\times d_l}$, with $N$ denoting the length of word embeddings. At each layer, the embedding $W_{i-1}$ is sent into the $i_{th}$ transformer layer $\mathcal{L}_{i}$ of the text encoder as:
\begin{equation}
\label{e3}
          [W_{i}] = \mathcal{L}_i([W_{i-1}]), i\in [0,\dots, K-1].
\end{equation} 
The final text representation $z$ is obtained by projecting the last token $w_{K-1}^N$ of the last transformer layer to the common V-L latent embedding space via a projection function $p_{l}$ as :
\begin{equation}
\label{e4}
        z = p_{L}(w_{K-1}^N).
\end{equation} 

\textbf{Zero-shot inference:} during inference, the inputs for the text encoder are hand-crafted prompts (e.g., 'A photo of a [class]) by replacing the placeholder [class] with the class name of label $y\in [1,\dots, C]$. Then the score for $j_{th}$ class is measured by calculating the similarity between outputs of the text encoder and image encoder via a cosine similarity function $sim()$ with a temperature parameter $\tau$ as:
\begin{equation}
\label{e5}
    p(y_j|x) = \frac{{\rm exp}(sim(x, z_j)/\tau)}{\sum_{i=1}^{C}{\rm exp}(sim(x, z_i))}.
\end{equation}
The class name corresponding to the highest score is adopted as the prediction result for the input image $I$.

The text prompts in the text encoder are usually hand-crafted which require labor-intensive manual search, and may not be optimal for the downstream task. Thus, recent methods propose to treat the prompts as textual embeddings and optimize them in the fine-tuning process. We next briefly introduce two typical soft-prompt-based methods.

\textbf{CoOp.} CoOp replaces the hand-crafted prompts in the text encoder as learnable soft prompts and directly updates them in the fine-tuning process. Specifically, CoOp introduces $M$ learnable prompt vectors \{$p_0$, $p_1$, \dots, $p_{M-1}$\}, and concatenate them with the token embedding $c_i$ of $i_{th}$ class as the text input embeddings: $t_i^{\rm CoOp}$ = \{$p_0$, $p_1$, \dots, $p_{M-1}$, $c_i$\}. These embeddings are then sent into the text encoder to obtain the final text representation. 


\textbf{MaPLe.} MaPLe argues that using prompting in a single branch of VLM may not be optimal since it doesn't allow adjusting representations of both branches to better match their output embeddings. Besides, only inserting prompts in the first layer may not be enough to encode beneficial information of various hierarchies. Thus, MaPLe proposes to insert prompts into both the vision and text branches up to $J$ layers to enable deep prompting. In each layer, the prompts of the image encoder are uni-directionally generated by the prompts of the text encoder.  

\subsection{Proposed Method}
\begin{figure*}[t]
    \centering
    \includegraphics[width=0.7\linewidth]{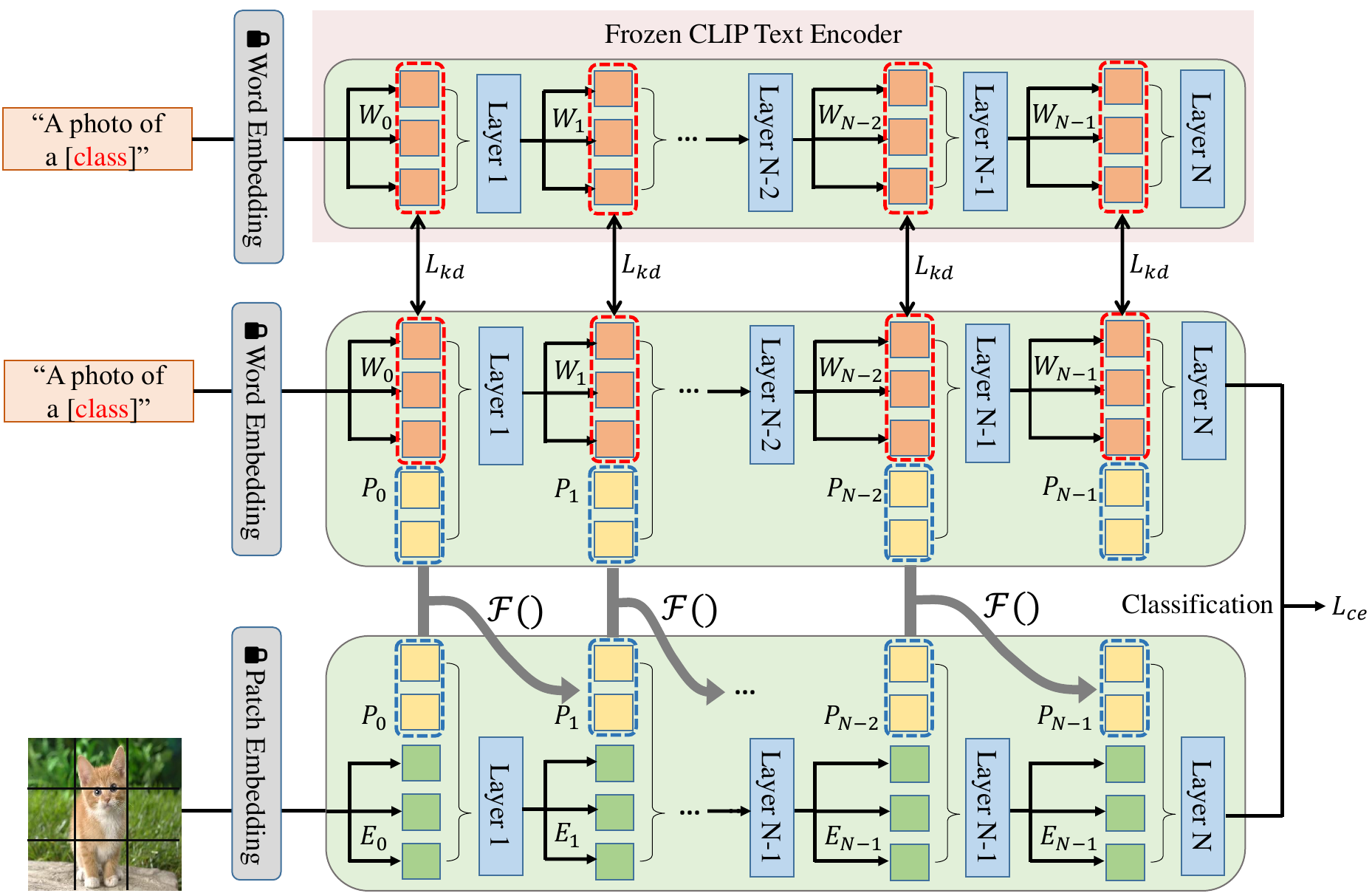} 
    \caption{The overview for COMMA. Here, $\mathcal{L}_{ce}$ denotes the cross-entropy loss and $\mathcal{L}_{kd}$ represents the knowledge distillation loss between two branches. COMMA generates the prompts of the vision branch based on preceding prompts of both branches to aggregate multi-modal beneficial information to guide their representation alignment. Besides, it let the learned prompts approximate the hand-crafted prompts in the pre-trained CLIP model to preserve generic knowledge.}
    \label{fig2}
  \end{figure*}

With considerable performance boost achieved by previous methods over downstream tasks compared to CLIP, they still have two major limitations. First, the prompts in the vision and text branches are usually separated or uni-directionally generated. Since the goal of VLM is to match the output embeddings of different modalities to describe their relationships, the prompts of both branches should be closely related to provide proper guidance to align the representations of both branches. Second, despite impressive performance boosts on the seen classes achieved by recent methods, they usually cause performance degeneration on the unseen classes compared to CLIP~\cite{yao2023visual,zhou2022conditional}, demonstrating worse generalization to novel concepts. This is harmful to real-life scenarios as a large number of novel classes may appear. The reason is that the generic knowledge is partly forgotten in the fine-tuning process. To handle the above limitations, we combine beneficial information from both branches to generate the prompts to well match their output embeddings. We also propose to preserve the generic representations of pretrained CLIP in the fine-tuning process to alleviate overfitting. An overview of our proposed method is given in fig.~\ref{fig2}. 

\subsubsection{Correlated prompt generation.} To better guide and align the representations of two branches, we present to compute prompts based on preceding prompts of both branches to aggregate beneficial multi-modal information. Especially, following previous methods~\cite{khattak2023maple}, we insert learnable prompts in both vision and text branches up to a specific depth $J$. Taking the vision branch as an example, the input embeddings are denoted as \{$P_0^v$, $c_0$, $E_0$\}, with $P$ representing the $M$-length learnable prompts. The calculation process of $i_{th}$ ($i\in [0,\dots, K-1]$) transformer layer $\mathcal{V}_i$ could be expressed as:

\begin{equation}
    \label{e6}
    [\_, c_{i}, E_{i}] = \mathcal{V}_i([P^v_{i-1}, c_{i-1}, E_{i-1}]).
\end{equation}

Instead of leaving the prompts in both branches separated or uni-directionally controlled, we leverage multi-modal information by computing the prompts in the image branch based on the preceding prompts of both branches. Specifically, for $i_{th}$ transformer layer $\mathcal{V}_i$, its prompts are dynamically generated by treating prompts $P_{i-1}^v$ of $(i-1)_{th}$ layer in the vision branch as a query, and the prompts $P_{i-1}^l$ of $(i-1)_{th}$ layer in the text branch as key and value. This procedure is expressed as :

\begin{equation}
    \label{e7}
    P_i^v = {\rm softmax}(\frac{P_{i-1}^v \cdot  P_{i-1}^l}{\sqrt{P}}) P_{i-1}^l.
\end{equation}
We perform aggregation along the token dimension. In this sense, the prompts in the vision encoder aggregate complementary information from the text branch to guide the alignment path of their representations. Practically, we only generate the prompts in the vision branch with guidance from the high-level semantics in the text branch, and leave the prompts in the text branch randomly initialized for back propagation to avoid hurting their high-level semantic representations.

\begin{figure}[t]
    \centering
    \includegraphics[width=\linewidth]{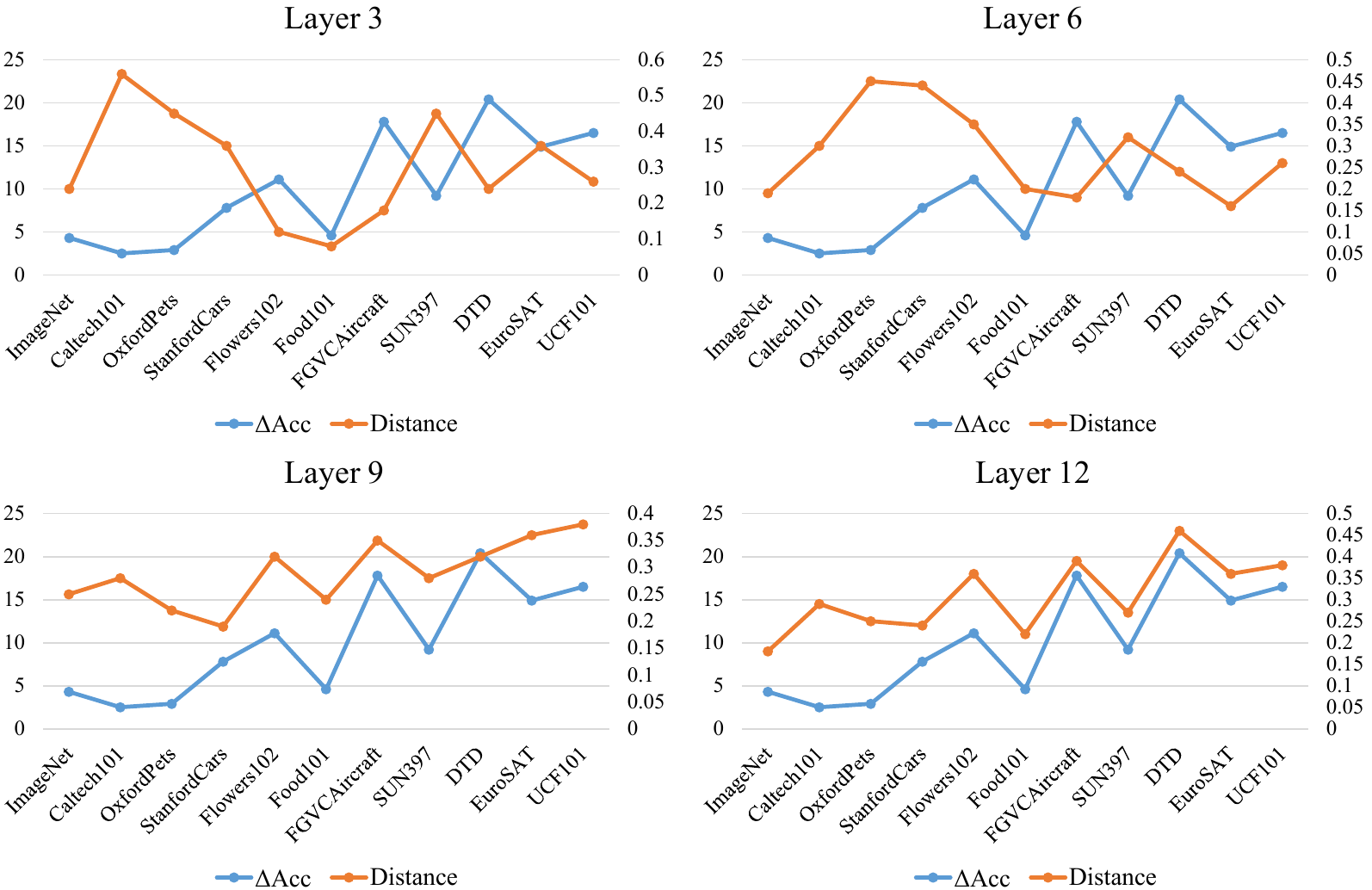} 
    \caption{Relationships concerning the degree of performance degradation $\Delta  {\rm Acc}$ with the distance between the learnable prompts in CoOp and the hand-crafted prompts in the pretrained CLIP across different layers over 11 datasets.}
    \label{fig3}
  \end{figure}
\subsubsection{Alleviating Forgetting Generic Knowledge.} Previous works~\cite{khattak2023maple,yao2023visual} have witnessed that the generic knowledge contained in pretrained CLIP models is easily forgotten in the fine-tuning process. We find that the similarity between the learned prompts and the hand-crafted prompts in pretrained CLIP is positively correlated with their performance on novel classes. Fig.~\ref{fig3} depicts the relationship concerning the distance between the learnable prompts in CoOp and the hand-crafted prompts in the pretrained CLIP with their performance gap $\Delta {\rm Acc}$ across different layers. It's observed that as the layers go deeper, the degree of performance degradation $\Delta {\rm Acc}$ is more consistent with the prompt embedding distance. Specifically, in the first few layers (e.g., Layer 3 \& 6) the correlations between $\Delta {\rm Acc}$ and the prompt distance are irregular, while in the last several layers (e.g., Layer 9 \& 12) the trends between $\Delta  {\rm Acc}$ and the prompt distance become more positively correlated. This indicates the distance between the learnable prompts and the hand-crafted prompts in the pretrained CLIP can be viewed as clear signs to indicate the model generalization performance over downstream tasks. Based on preceding observations, we propose to minimize the feature discrepancy between the learnable prompts and hand-crafted prompts of the pretrained CLIP in the last several $S$ layers, to boost the generalization performance on novel classes. Specifically, for the reciprocal $s_{th}$ layer, we maximize the feature similarity between the learnable prompts in the text branch of COMMA and the hand-crafted prompts in the text branch of the pretrained CLIP via a cosine similarity $sim()$ as:   

\begin{equation}
    \label{e8}
    \mathcal{L}_{\rm kd} = Sim(P_s^l, P_s^{\rm CLIP})
\end{equation}

Overall, we minimize the cross-entropy loss as well as the feature discrepancy loss with the weight $\lambda$ over the reciprocal $S$ layers to train our COMMA as :
\begin{equation}
    \label{e9}
    \mathcal{L}_{\rm Total} = \mathcal{L}_{\rm ce} + \lambda\sum_{i=0}^{S}(1-L_{\rm kd}^i). 
\end{equation}
\section{Experiments}
\subsection{Benchmark Setting}
\subsubsection{Base-to-Novel Generalization:} The datasets are split into base and novel classes to evaluate the model in a zero-shot manner. The model is trained on the base classes in a few-shot setting and evaluated on base and novel classes.
\subsubsection{Cross-dataset Evaluation:} To demonstrate the ability of our model in cross-dataset transfer, we train our model on the ImageNet dataset in a few-shot manner, and directly evaluate it on other datasets without further fine-tuning.
\subsubsection{Domain Generalization:} To evaluate the robustness of our model over out-of-distribution data, we directly test our ImageNet-trained model on four other ImageNet datasets which contain different types of domain shifts.
\subsubsection{Datasets}: For \textbf{base-to-novel generalization} and \textbf{cross-dataset evaluation}, we follow previous methods~\cite{khattak2023maple,yao2023visual} to evaluate the performance of our method on 11 image classification datasets, including two generic-objects datasets, ImageNet~\cite{deng2009imagenet} and Caltech101~\cite{fei2004learning}; five fine-grained datasets, OxfordPets~\cite{parkhi2012cats}, StanfordCars~\cite{krause20133d}, Flowers102~
\cite{nilsback2008automated}, Food101~\cite{bossard2014food}, and FGVCAircraft~\cite{maji2013fine}; a scene recognition dataset SUN397 ~\cite{xiao2010sun}; an action recognition dataset UCF101~\cite{soomro2012ucf101}; a texture dataset DTD~\cite{cimpoi2014describing} and a satelliteimage dataset EuroSAT~\cite{helber2019eurosat}. For \textbf{domain generalization}, we use ImageNet as the source dataset and its four variants as target datasets including ImageNetV2~\cite{recht2019imagenet}, ImageNet-Sketch~\cite{wang2019learning}, ImageNet-A~\cite{hendrycks2021natural} and ImageNet-R~\cite{hendrycks2021many}.

\subsubsection{Implementation Details} For all experiments, we use the pretrained ViT-B/16 CLIP model by default with $d_l$ = 512, $d_l$ = 768. We use a 16-shot training strategy in all experiments by default which randomly samples 16 shots for each class. Following previous methods~\cite{khattak2023maple}, we set prompt depth $J$ to 9 and the language and vision prompt lengths to 2. We train our models for 5 epochs with a batch-size of 4 and a learning rate of 0.0035 with the SGD optimizer. We use the pretrained CLIP word embeddings of the template 'a photo of a [category]' to initialize the language prompts of the first layer $P_0$, and randomly initialize the prompts of the subsequent layers with a normal distribution. For \textbf{base-to-novel generalization}, we report base and novel class accuracies and their harmonic mean (HM) averaged over 3 runs. For \textbf{cross-dataset evaluation} and \textbf{domain generalization}, we train our model on the ImageNet dataset as a source model for 2 epochs with a learning rate of 0.0026, and set the prompt depth $J$ as 3.

\begin{table*}[h!]
   
    \begin{subtable}{.33\linewidth}
    \centering 
    \begin{tabular}{l|cc|c}
    \toprule & Base & New & HM \\
    \midrule CLIP & 69.34 & 74.22 & 71.70 \\
    CoOp & \textbf{82.63} & 67.99 & 74.60 \\
    CoCoOp & 80.47 & 71.69 & 75.83 \\
    KgCoOp & 80.73 & 73.60 & 77.00 \\
    MaPLe & 82.28 & 75.14& 78.55 \\
    \midrule 
    COMMA & 82.42 & \textbf{75.87}& \textbf{79.04} \\
    \hline
    \end{tabular}
    \caption{Average over 11 datasets.}
    \end{subtable}%
    \begin{subtable}{.33\linewidth}
    \centering 
        \begin{tabular}{l|cc|c}
        \toprule & Base & New & HM \\
        \midrule CLIP & 72.43 & 68.14 & 70.22 \\
        CoOp & \textbf{76.46} & 66.31 & 71.02 \\
        CoCoOp & 75.98 & 70.43 & 73.10 \\
        KgCoOp & 75.83 & 69.96 & 72.78 \\
        MaPLe & 76.66 & 70.54 & 73.47 \\
        \midrule 
        COMMA & 76.04 & \textbf{70.89} & \textbf{73.86} \\
        \hline
        \end{tabular}
    \caption{ImageNet.}
    \end{subtable}%
    \begin{subtable}{.33\linewidth}
    \centering 
    \begin{tabular}{l|cc|c}
    \toprule & Base & New & HM \\
    \midrule CLIP & 96.84 & 94.00 & 95.40 \\
    CoOp & \textbf{98.11} & 93.52 & 95.76 \\
    CoCoOp & 97.96 & 93.81 & 95.84 \\
    KgCoOp & 97.72 & 94.39 & 96.03 \\
    MaPLe & 97.74 & 94.36 & 96.02 \\
    \midrule 
    COMMA & 97.94 & \textbf{94.56} & \textbf{96.50} \\
    \hline
    \end{tabular}
    \caption{Caltech101.}
    \end{subtable}%

    \begin{subtable}{.33\linewidth}
    \centering 
    \begin{tabular}{l|cc|c}
        \toprule & Base & New & HM \\
        \midrule CLIP & 91.17 & 97.26 & 94.12 \\
        CoOp & 94.24 & 96.66 & 95.43 \\
        CoCoOp & 95.20 & 97.69 & 96.43 \\
        KgCoOp & 94.65 & 97.76 & 96.18 \\
        MaPLe & 95.43 & 97.76& 96.58 \\
        \midrule 
        COMMA & \textbf{95.62} & \textbf{97.84} & \textbf{96.72} \\
        \hline
        \end{tabular}
        \caption{OxfordPets.}
    \end{subtable}%
    \begin{subtable}{.33\linewidth}
    \centering 
    \begin{tabular}{l|cc|c}
        \toprule & Base & New & HM \\
        \midrule CLIP & 63.37 & 74.89 & 68.65 \\
        CoOp & \textbf{76.20} & 69.14 & 72.49 \\
        CoCoOp & 70.49 & 73.59 & 72.01 \\
        KgCoOp & 71.76 & 75.04 & 73.36 \\
        MaPLe & 72.94 & 74.00 & 73.47 \\
        \midrule
        COMMA & 73.48 & \textbf{74.91} & \textbf{73.96} \\
        \hline
        \end{tabular}
        \caption{StanfordCars.}
    \end{subtable}%
    \begin{subtable}{.33\linewidth}
    \centering 
    \begin{tabular}{l|cc|c}
    \toprule & Base & New & HM \\
    \midrule CLIP & 72.08 & \textbf{77.80} & 74.83 \\
    CoOp & \textbf{97.63} & 69.55 & 81.23 \\
    CoCoOp & 94.87 & 71.75 & 81.71 \\
    KgCoOp & 95.00 & 74.73 & 83.65\\
    MaPLe & 95.92 & 72.46 & 82.56\\
    \midrule 
    COMMA & 94.86 & 75.13 & \textbf{83.88}\\
    \hline
    \end{tabular}
    \caption{Flowers102.}
    \end{subtable}%

    \begin{subtable}{.33\linewidth}
    \centering 
    \begin{tabular}{l|cc|c}
        \toprule & Base & New & HM \\
        \midrule CLIP & 90.10 & 91.22 & 90.66 \\
        CoOp & 89.44 & 87.50 & 88.46 \\
        CoCoOp & 90.70 & 91.29 & 90.99 \\
        KgCoOp & 90.50 & 91.70 & 91.09 \\
        MaPLe & \textbf{90.71} & 92.05 & 91.38 \\
        \midrule 
        COMMA & 90.42 & \textbf{92.74} & \textbf{91.84} \\
        \hline
        \end{tabular}
        \caption{Food101.}
    \end{subtable}%
    \begin{subtable}{.33\linewidth}
    \centering
    \begin{tabular}{l|cc|c}
        \toprule & Base & New & HM \\
        \midrule CLIP & 27.19 & \textbf{36.29} & 31.09 \\
        CoOp & \textbf{39.24} & 30.49 & 34.30 \\
        CoCoOp & 33.41 & 23.71 & 27.74 \\
        KgCoOp & 36.21 & 33.55 & 34.83 \\
        MaPLe & 37.44 & 35.61 & \textbf{36.50} \\
        \midrule 
        COMMA & 36.47 & 34.23 & 35.84 \\
        \hline
        \end{tabular}
        \caption{FGVCAircraft.}
    \end{subtable}%
    \begin{subtable}{.33\linewidth}
    \centering
    \begin{tabular}{l|cc|c}
        \toprule & Base & New & HM \\
        \midrule CLIP & 69.36 & 75.35 & 72.23 \\
        CoOp & 80.85 & 68.34 & 74.07 \\
        CoCoOp & 79.74 & 76.86 & 78.27 \\
        KgCoOp & 80.29 & 76.53 & 78.36 \\
        MaPLe & 80.82 & 78.70 & 79.75 \\
        \midrule 
        COMMA & \textbf{80.94} & \textbf{79.32} & \textbf{80.86} \\
        \hline
        \end{tabular}
        \caption{SUN397.}
    \end{subtable}%

    \begin{subtable}{.33\linewidth}
    \centering
    \begin{tabular}{l|cc|c}
        \toprule & Base & New & HM \\
        \midrule CLIP & 53.24 & \textbf{59.90} & 56.37 \\
        CoOp & 80.17 & 47.54 & 59.68 \\
        CoCoOp & 77.01 & 56.00 & 64.85 \\
        KgCoOp & 77.55 & 54.99 & 64.35 \\
        MaPLe & 80.36 & 59.18 & 68.16 \\
        \midrule
        COMMA & \textbf{81.04} & 58.62 & \textbf{68.32} \\
        \hline
        \end{tabular}
        \caption{DTD.}
    \end{subtable}%
    \begin{subtable}{.33\linewidth}
    \centering
    \begin{tabular}{l|cc|c}
        \toprule & Base & New & HM \\
        \midrule CLIP & 56.48 & 64.05 & 60.03 \\
        CoOp & 91.54 & 54.44 & 68.27 \\
        CoCoOp & 87.49 & 60.04 & 71.21 \\
        KgCoOp & 85.64 & 64.34 & 73.48 \\
        MaPLe & \textbf{94.07} & 73.23 & 82.35 \\
        \midrule 
        COMMA & 93.56 & \textbf{74.26} & \textbf{83.42} \\
        \hline
        \end{tabular}
        \caption{EuroSAT.} 
    \end{subtable}%
    \begin{subtable}{.33\linewidth}
    \centering
    \begin{tabular}{l|cc|c}
    \toprule 
    & Base & New & HM \\
    \midrule
    CLIP & 70.53 & 77.50 & 73.85 \\
    CoOp & \textbf{85.14} & 64.47 & 73.37 \\
    CoCoOp & 82.33 & 73.45 & 77.64 \\
    KgCoOp & 82.89 & 76.67 & 79.65 \\
    MaPLe & 83.00 & 78.66 & 80.77 \\
    \midrule 
    COMMA & 84.06 & \textbf{80.56} & \textbf{81.84} \\
    \hline
    \end{tabular}
    \caption{UCF101.}
    \end{subtable}%

    \caption{Comparison with recent methods in the base-to-new generalization setting. 'HM' denotes Harmonic mean.} 
    \label{tab1}
\end{table*}

\subsection{Base-to-Novel Generalization}

We split each dataset into two disjoint groups: base classes (Base) and new classes (New). The model is trained on the base classes and directly evaluated on the unseen new classes to validate its generalizability. We compare our COMMA with recent methods in tab.~\ref{tab1}. Totally, COMMA achieves improved performance in 9/11 datasets upon new classes and higher harmonic mean accuracies over 10/11 datasets compared to state-of-the-art methods, demonstrating better generalizability over novel concepts. Specifically, previous methods like CoOp, CoCoOp and KgCoOp usually achieve large improvements upon base classes compared to CLIP. However, they often own worse performance on the novel classes. This is because they easily overfit the training data and lack generalization on unseen data. As a strong competitor, MaPLe introduces multi-modal prompt learning to alleviate this issue and improves a lot over new classes compared to previous methods. Our COMMA not only demonstrates much superior performance over all base classes compared to CLIP, but also shows stronger accuracy upon most (9/11) new classes with impressive generalizability, with a higher averaged harmonic mean accuracy upon all datasets. It's worth noting that compared to previous methods, our COMMA just obtains higher performance over 3/11 datasets on base classes, but achieves better accuracy over 9/11 datasets on novel classes.

\subsection{Cross-Dataset Transfer}
We test cross-dataset generalization of COMMA in tab.~\ref{tab2}. The model is trained on the ImageNet dataset and directly evaluated on the remaining 10 datasets. It's observed that COMMA achieves competitive performance with other methods on the source ImageNet dataset, but achieves much stronger performance over the target datasets. Especially, COMMA outperforms CoOp and MaPLe over 9/10 datasets and beats CoCoOp over all datasets. Overall, COMMA achieves the highest averaged performance over all 10 datasets, with better generalizability over downstream tasks.  

\begin{table*}[t]
    \setlength\tabcolsep{1pt}
    \centering
    \begin{tabular}{ccccccccccccc}
      \toprule
      & Source & \multicolumn{11}{c}{Target} \\
      \cmidrule(r){2-2}
      \cmidrule(r){3-13}
      & ImageNet & Caltech101 & OxfordPets & S-Cars & Flowers102 & Food101 & Aircraft & SUN397 & DTD & EuroSAT & UCF101 & Average \\
      \midrule 
      CoOp & \textbf{71.51} & 93.70 & 89.14 & 64.51 & 68.71 & 85.30 & 18.47 & 64.15 & 41.92 & 46.39 & 66.55 & 63.88 \\
      CoCoOp & 71.02 & 94.43 & 90.14 & 65.32 & 71.88 & 86.06 & 22.94 & 67.36 & 45.73 & 45.37 & 68.21 & 65.74 \\
      MaPLe & 70.72 & 93.53 & 90.49 & 65.57 & 72.23 & \textbf{86.20} & 24.74 & 67.01 & 46.49 & 48.06 & 68.69 & 66.30 \\
      \midrule 
      COMMA & 71.22 & \textbf{93.84} & \textbf{90.78} & \textbf{66.36} & \textbf{73.14} & 85.87& \textbf{25.14} & \textbf{67.56} & \textbf{46.52} & \textbf{48.85} & \textbf{68.71} & \textbf{66.84} \\
      \bottomrule 
    \end{tabular}
    \caption{Comparison of COMMA with recent methods on the cross-dataset evaluation setting. }
    \label{tab2}
  \end{table*} 

\begin{table}[t]
    \setlength\tabcolsep{1pt}
    \centering
    \begin{tabular}{cccccc}
      \toprule
      & Source & \multicolumn{4}{c}{Target} \\
      \cmidrule(r){2-2}
      \cmidrule(r){3-6}
      & ImgNet & ImgNetV2 & ImgNet-S & ImgNet-A & ImgNet-R \\
      \midrule 
      CLIP & 66.73 & 60.83 & 46.15 & 47.77 & 73.96 \\
      CoOp & \textbf{71.51} & 64.20 & 47.99 & 49.71 & 75.21 \\
      CoCoOp & 71.02 & 64.07 & 48.75 & 50.63 & 76.18 \\ 
      MaPLe & 70.72 & 64.07 & 49.15 & 50.90 & 76.98 \\
      KgCoOp & 71.20 & 64.10 & 48.97 & 50.69 & 76.70 \\
      \midrule
      COMMA & 71.22 & \textbf{64.84} & \textbf{49.65} & \textbf{51.64} & \textbf{77.56} \\
      \bottomrule 
    \end{tabular}
    \caption{Comparison of COMMA with existing approaches in the domain generalization setting. }
    \label{tab3}
  \end{table}
  
\subsection{Domain Generalization}
We train our model on the source ImageNet dataset, and directly evaluate it on four out-of-distribution datasets to test its generalizability. The results are shown in tab.~\ref{tab3}. Our COMMA achieves competitive performance on the source ImageNet dataset against other methods, and consistently outperforms other methods across all other datasets. This indicates that enhancing the correlations of multi-modal prompts and injecting generic knowledge could enhance the generalization and robustness of VLMs like CLIP.

\subsection{Ablation Study}

\subsubsection{Effectiveness of proposed components.} We validate the effects of the proposed two components, i.e., correlated prompt generalization and generic knowledge transfer, in tab.~\ref{tab4}. It's observed adding correlate prompts and knowledge transfer could bring +2.16\% \& 0.98\% averaged performance boost over all 11 datasets. Combining both further leads to a +3.18\% accuracy boost with absolute 79.04\% accuracy.

\begin{table}[t]
\setlength\tabcolsep{2pt}
\centering
\begin{tabular}{cc}
\hline 
Configurations & Accuracy(\%) \\
\hline
 - &  75.86 \\
 w/ correlated prompts  & 78.02 \\
 w/ knowledge transfer & 76.84 \\
 COMMA  & \textbf{79.04} \\
\hline 
\end{tabular}
\caption{Effectiveness of each component in COMMA. }
\label{tab4}
\end{table} 

\begin{figure}[h!]
    \centering
    \includegraphics[width=0.8\linewidth]{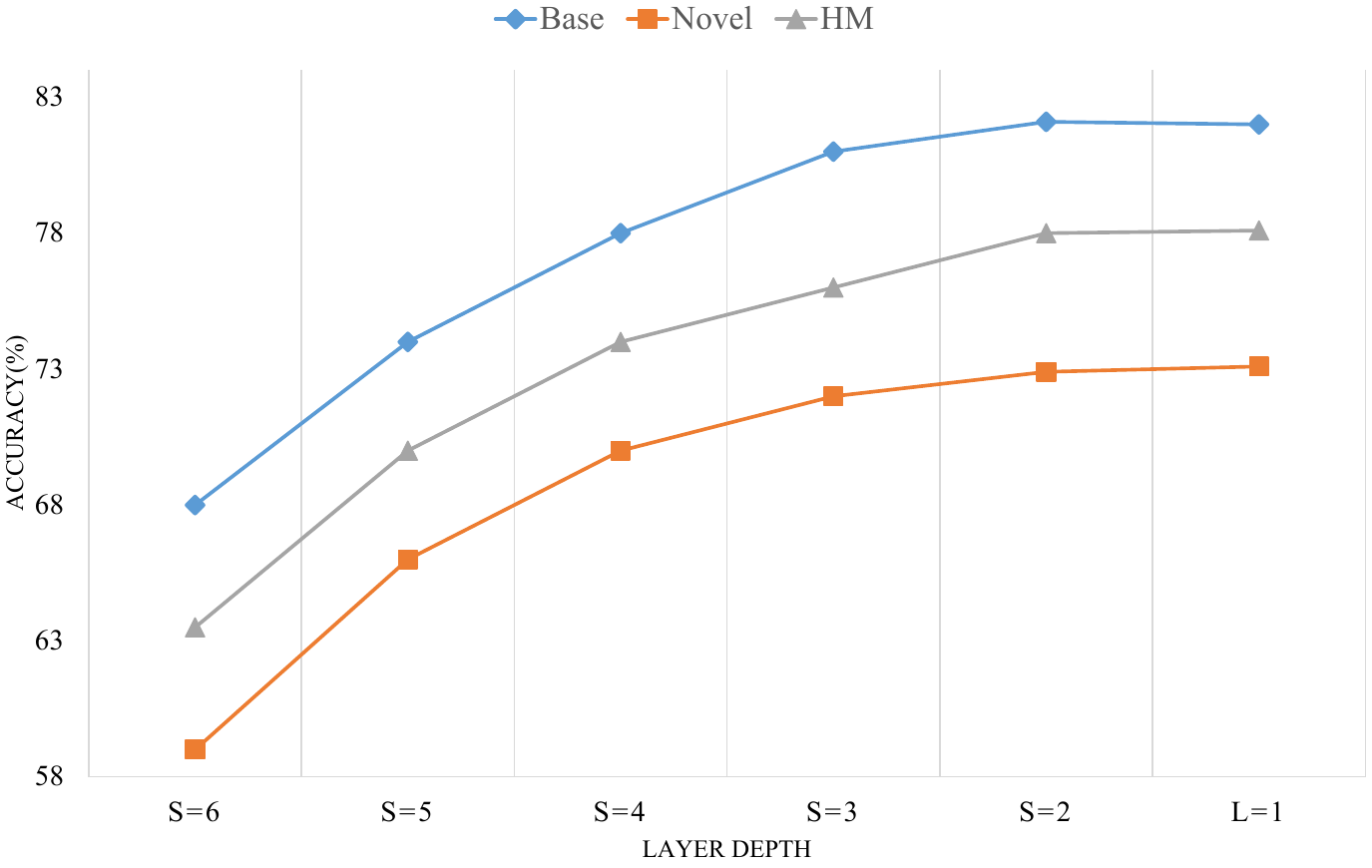} 
    \caption{Relationships concerning the base-class, novel-class and harmonic mean accuracy with the number of reciprocal layers ($S$). Accuracies are averaged over 11 datasets.}
    \label{fig4}
  \end{figure}  

\begin{table}[t]
    \centering
    \begin{tabular}{cccccc}
    \hline 
    \multirow{2}{*}{Method} & \multirow{2}{*}{Params } & \multicolumn{2}{c}{FPS (with BS)} & \multirow{2}{*}{HM}\\
    & & 1 & 16 & \\
    \hline 
    CoOp & 2048  & 9.4 & 147.6 & 71.66 \\
    CoCoOp & 35360 & 45.2 & 726.2 & 75.83 \\
    KgCoOp & 2048  & 40.1 & 642.3 & 77.00 \\
    MaPLe & 3.55M & 39.8 & 639.8 & 78.55 \\
    \hline
    COMMA & 4.87M & 39.6 & 637.8 & \textbf{79.04} \\
    \hline
    \end{tabular}
    \caption{Comparison of prompting efficiency with other methods over 11 datasets. 'BS' denotes 'Batch Size'. }
    \label{tab5}
    \end{table} 

\begin{table}[t]
    \setlength\tabcolsep{3pt}
    \centering
    \begin{tabular}{ccccccc}
    \hline 
    $\lambda$ & 0.5 & 1.0 & 1.5 & 2.0 & 4.0 & 8.0 \\
    \hline 
    Accuracy & 78.68  & \textbf{79.04} & 78.74 & 78.62 & 78.21 & 78.02 \\
    \hline
    \end{tabular}
    \caption{Effects for the weight $\lambda$ over 11 datasets. }
    \label{tab6}
    \end{table}    
\subsubsection{How many layers to adopt knowledge transfer.} We use the reciprocal $S$ layers to transfer generic knowledge from the hand-crafted prompts in CLIP to the learnable prompts in COMMA. Fig.~\ref{fig4} plots the accuracy variation of base class, novel class and harmonic mean by changing $S$. It's observed that the accuracies consistently rise as $S$ decrease, which reach the peak when $S$=2. This indicates that the prompts embeddings in the last several layers contain more generic semantic information, which can better help the generalization performance over downstream tasks. We thus set $S$=2.

\subsubsection{Prompting efficiency.} We compare the efficiency of COMMA with recent methods concerning parameters and frames per second (FPS) in tab.~\ref{tab5}. It's noticed the CoCoOp has the lowest FPS due to its instance-conditioned design. Its FPS decreases as the batch size rises, which greatly hinders it application in real life. With improved accuracy compared to previous methods (CoOp, CoCoOp and KgCoOp), MaPLe increases the required parameters from 2048 to 3.55M. Compared to other methods, our COMMA owns slightly increased parameters with comparable FPS and the highest averaged performance over all 11 datasets, demonstrating better accuracy with competitive computational costs.
 
\subsubsection{The choice of $\lambda$.} We test the choice for the weight $\lambda$ of the knowledge transfer loss $\mathcal{L}_{kg}$ in tab.~\ref{tab6}. As $\lambda$ rises, the accuracy increases and reaches the peak when $\lambda$=1.0. We set $\lambda$=1.0 by default.

\subsubsection{The choices of prompt depth $J$ and prompt length $P$.}
We ablate the choices of prompt depth $J$ and prompt length $P$ in tab.~\ref{tab7}. It's observed that $J=9$ and $P=2$ could offer the best results.

\begin{table}[h]
    \setlength\tabcolsep{3pt}
    \centering
    \begin{tabular}{cccc|cccc}
    \hline
    $J$=3 &  $J$=6 & $J$=9 & $J$=12 & $P$=1 & $P$=2 & $P$=3 & $P$=4\\
    \hline 
    77.98 & 78.56 & \textbf{79.04} & 78.74 & 78.64 & \textbf{79.04} & 78.75 & 78.54 \\
    \hline 
    \end{tabular}
    \caption{Ablations for prompt depth $J$ and prompt length $P$. }
    \label{tab7}
    \end{table} 

\section{Conclusion}
Adapting large VLMs for downstream tasks is challenging due to the large scale of optimized parameters and limited size of downstream data. Prompt learning is an efficient promising approach to adapt VLMs to novel concepts. To increase the generalization performance of VLMs, we propose to enhance the correlations of multi-modal prompts and preserve generic knowledge during the fine-tuning process. Experimental results show that our model is both parameter-efficient and robust across a series of downstream tasks.

\section{Acknowledgments}
This work is supported by National Key Research and Development Program of China (2020YFC1522700) and National Natural Science Foundation of China (Project No. 62072334).
\bibliography{ref}
\end{document}